\documentclass[journal]{IEEEtran}
\IEEEoverridecommandlockouts
\usepackage{amsmath,amsfonts}
\usepackage{algorithmic}
\usepackage{algorithm}
\usepackage{array}
\usepackage[caption=false,font=normalsize,labelfont=sf,textfont=sf]{subfig}
\usepackage{textcomp}
\usepackage{stfloats}
\usepackage{float}
\usepackage{placeins}
\usepackage{url}
\usepackage{verbatim}
\usepackage{graphicx}
\usepackage{cite}
\usepackage{soul}
\usepackage[dvipsnames]{xcolor}
\newcommand{\realnumbers}{\mathbb{R}}

\begin{document}

\title{VeMo: A Lightweight Data-Driven Approach \\to Model Vehicle Dynamics}

\author{
Girolamo Oddo, Roberto Nuca$^{*}$, Matteo Parsani\\
King Abdullah University of Science and Technology\\
\thanks{$^*$ Corresponding author (roberto.nuca@kaust.edu.sa).}
}



\maketitle

\begin{abstract}
Developing a dynamic model for a high-performance vehicle is a complex problem that requires extensive structural information about the system under analysis. This information is often unavailable to those who did not design the vehicle and represents a typical issue in autonomous driving applications, which are frequently developed on top of existing vehicles; therefore, vehicle models are developed under conditions of information scarcity. This paper proposes a lightweight encoder-decoder model based on Gate Recurrent Unit layers to correlate the vehicle's future state with its past states, measured onboard, and control actions the driver performs. The results demonstrate that the model achieves a maximum mean relative error below 2.6\% in extreme dynamic conditions. 
It also shows good robustness when subject to noisy input data across the interested frequency components. Furthermore, being entirely data-driven and free from physical constraints, the model exhibits physical consistency in the output signals, such as longitudinal and lateral accelerations, yaw rate, and the vehicle's longitudinal velocity.
\end{abstract}


\section{Introduction}
\IEEEPARstart{I}{n} the automotive sector developing a representative vehicle dynamics model is a complex and multifaceted challenge \cite{jazar2017vehicle,milliken1996vehicle,guiggiani2023vehicle}. Numerous nonlinear factors influence vehicle dynamics, including tire characteristics, suspension geometry, aerodynamics, drivetrain effects, and external environmental factors, such as road surface grip conditions and climatic effects (e.g., wind). Accurately capturing these effects in a computational model requires high-fidelity multibody simulation software and a profound understanding of the vehicle system. These resources and insights are typically known only by the vehicle manufacturer and are inaccessible to third parties.

This limitation is particularly evident in the context of autonomous or advanced driver assistance systems (ADAS) , which are frequently developed as aftermarket solutions on top of existing vehicle platforms \cite{autonomous_urban_driving}. The lack of access to critical vehicle parameters restricts the ability to construct precise physics-based models, resulting in the operational use of highly simplified models even in applications involving high-performance vehicles or in extreme operating conditions such as drifting \cite{vehicle_dynamics_survey,drifting_trajectories}. Furthermore, nonlinear dynamic effects, crucial in high-performance applications, are challenging to approximate with simplified methods.

In this context, data-driven approaches have emerged as a valid modeling alternative \cite{b_sindy, dd_ssm}. By leveraging onboard data sensors and modern machine learning techniques, it becomes possible to infer complex system dynamics directly from empirical observations, bypassing the problem of obtaining accurate models in conditions of limited or complete lack of constructive information about the vehicle system under examination.

This work is geared towards developing an accurate, robust, light-weight, and almost real-time vehicle dynamic model for intelligent and automated vehicles~\cite{vehicle_dynamics_survey}. To this end, we propose \textit{Vehicle Modeller} (VeMo, see \cite{VeMoRepository} for the source code), a neural model based on an encoder-decoder architecture, using Gate Recurrent Unit (GRU) layers \cite{cho2014rnn}, with a shared encoder and dedicated decoders for each output. This approach preserves the dynamic coupling of the input signals while enabling each decoder to specialize in decoding its target signal. Although this technique of branching the decoder is known in the fields of computer vision and time series applications (see \cite{MultiBranch1, MultiBranch2, MultiBranch3}), it appears to be absent in the context of vehicle dynamics modeling using neural network-based approaches.    

The model is tasked with predicting the next vehicle state, expressed in terms of longitudinal and lateral accelerations, yaw rate, and longitudinal velocity based on measured past states, the driver's control actions, and their history. These quantities are selected as state variables because they correspond to the outputs of inertial measurement units (IMU) and tone wheels, sensors commonly available onboard, even in low-cost prototypes. Thus, this approach is readily integrable with control systems, which frequently rely on these measurements \cite{IMUcontroller1, IMUcontroller2}. Notably, the proposed procedure is not limited to the selected state variables of this work; it can accommodate other onboard-measured quantities. Including yaw rate and velocity alongside accelerations underscores the model's flexibility, as it is not constrained to a specific order of differentiation for state variables—such as velocities, accelerations, or jerks—enabling diverse representations of vehicle states. 
Moreover, the procedure readily extends across vehicle platforms (wheeled, tracked, and aerial) without altering its core formulation.

To validate the model, we use a GT3-class\footnote{The GT3-class or group GT3, known technically as cup grand touring cars and commonly referred to as simply GT3, is a set of regulations maintained by the Fédération Internationale de l'Automobile (FIA) for grand tourer racing cars designed for use in various auto racing series throughout the world.} racing vehicle, operating under conditions beyond the linear dynamic range to explore the vehicle dynamics in extreme scenarios. This choice allows us to test the accuracy and robustness of the proposed models thoroughly.

To evaluate the proposed approach and its robustness, we compare the logged signals and those obtained from the model in one-step prediction mode with different levels of high-frequency filtering, specifically analyzing the cases of $45\,Hz$, $25\,Hz$, $5\,Hz$, and $0.5\,Hz$. These are frequencies at which the data are filtered to train the models. The one-step prediction mode also provides insight into the performance's sensitivity to noise in training. Furthermore,  to assess the model's performance under more intense noise disturbances than those used during training, the neural models are evaluated on inputs filtered at cutoff frequencies greater or equal to those used in the training phase of each respective model: $45\,Hz$, $25\,Hz$, $15\,Hz$, $5\,Hz$, and $1\,Hz$. The latter tests also help to understand how noise influences the models' performance and their response to unmodeled noise, a must-know information for safety-critical applications. This challenge is frequently addressed in developing controllers based on predictive models, such as model predictive control (MPC), as noted in \cite{pinn_mpc_w_noise}, since noise is an unavoidable factor in such applications. Furthermore it is well-established that noise can degrade the representational power of models trained through supervised learning \cite{noise1,noise2,noise3}.

The numerical results of the tests show that the VeMo model delivers excellent relative error and physical coherence performance, even without explicitly enforced physical constraints. Although the models are robust to noise, the study of the noise influence on VeMo's performance highlights the importance of using filtering frequencies similar to those applied during the model's training.


\section{Related Work}

Numerous studies have recently emerged on the topic of vehicle dynamics modeling in contexts where purely physical models are not achievable due to limitations in the available system information. These studies have reported diverse approaches, particularly regarding the presence or absence of physical constraints or models in neural systems.

In vehicle dynamics modeling through neural approaches, the growing interest in machine learning has led to a surge in recent works. Specifically, with the advent of the physics-informed neural network (PINN) paradigm \cite{pinn}, many researchers have focused on its application to vehicles. For example, in \cite{deepdyn} and \cite{deepdyn_pred}, the authors propose a two-stage system: the first part employs a neural network capable of identifying the parameters of the vehicle under study, the second part characterizes a single-track dynamic model for simulation purposes. Between these two stages, \cite{deepdyn} incorporates a ``guard" layer to ensure the physical consistency of the estimated parameters.

Other researchers present hybrid approaches, where a neural model operates downstream of the system as a corrector for the physical model, using various network architectures. Examples include approaches using fully connected layers \cite{spielberg}, LSTMs \cite{phy_dl}, and more complex architectures like attention-GRU \cite{att_gru}. The latter approach focuses on scenarios with limited data.

Beyond these cases of networks informed or supported by physical constraints, there are also purely neural approaches known as ``End-to-End," where the network is required to map inputs directly to outputs without additional external support. For clarity, the term ``End-to-End," in the context of autonomous driving, is also used in the literature with a different meaning, referring not to vehicle dynamics modeling but to the direct control actions derived from sensors and the perceived environment, as reported in \cite{e2e_autonomous_survey}. The primary work in vehicle dynamics that adopted this strategy, introducing GRU layers, is \cite{hermansdorfer}, where a recursive approach is presented to simulate vehicle dynamics by providing past states and control actions to predict the next state.
Differing from what is proposed in this work in terms of architecture, they propose a purely sequential approach with GRU units without branches, while in this work we focus on an encoder-decoder structure.

Further work has been conducted using this ``End-to-End" approach, exploring the application of fully connected networks \cite{e2efnn} or LSTM cells \cite{e2elstm} to predict lateral acceleration and lateral velocity, respectively, showing good performance. Additional work under this paradigm is reported in \cite{e2elonglat}, employing a fully connected network approach to predict longitudinal acceleration and yaw rate using two separate networks.

Currently, the literature does not identify an optimal approach across this spectrum of emerging variants, making it particularly interesting to explore cases with varying degrees of, or entirely without, physical constraints coupled with neural models, as done in this work. This allows for the provision of diverse solutions based on the level of knowledge available about the target system to be modeled.

\section{Problem Statement}
The problem addressed in this study can be mathematically described using a uniform time-step discretization $t_n=n~\Delta t$, where $n=0,1,2,\dots$ is the subscript addressing the time index and $\Delta t$ is the (constant) time step. The value of $\Delta t$ is defined by the sampling rate of the sensors, set to $100~Hz$ in this work; further details are provided in Section \ref{sec:data}. The problem is stated as searching for a neural network model $\mathcal{F}$ that computes the new state $\mathbf{x}_{n+1}$ as a function of the immediately past $k$ states and control actions, i.e.,
\begin{equation}
    \label{eq:discretedysys}
    \mathbf{x}_{n+1} = \mathcal{F}\left( \mathbf{x}_{n-k+1},\dots,\mathbf{x}_n ; \mathbf{u}_{n-k+1},\dots,\mathbf{u}_n \right),
\end{equation}
where $\mathbf{x}_{j}\in\realnumbers^4$ is the vehicle state vector composed by the longitudinal acceleration, $a_x$, the lateral acceleration, $a_y$, the yaw rate $\dot{\theta}$, and the longitudinal velocity $v_x$, i.e.
$$\mathbf{x} =
    \left[
        \begin{array}{c}
            a_x \\
            a_y \\[4pt]
            \dot\theta \\
            v_x\\
        \end{array}
    \right],
$$
whereas $\mathbf{u}_j \in \realnumbers^4$ is the control action vector defined as
$$
\mathbf{u} =
    \left[
        \begin{array}{c}
            u_t \\
            u_b \\
            u_s \\
            u_g
        \end{array}
    \right], 
$$
where $u_t$, $u_b$, $u_s$, and $u_g$ are the throttle percentage, the brake percentage, the steering angle, and the gear, respectively. Both past vehicle states and control actions (i.e., $\mathbf{x}_{j}$ and $\mathbf{u}_j$ with $j\leq~n$) are considered to capture possible delays in the effect of controls on the vehicle state.

The vehicle states are always referenced to a right-handed frame fixed to the vehicle, with $x$, $y$ forming a plane parallel to the ground, as specified in the standard ISO 8855:2011(E) system for ground vehicles. Therefore, in equation \eqref{eq:discretedysys}, the neural network $\mathcal{F}$ represents a black-box transfer function that relates past states and control actions to the future state of the vehicle. The vehicle state and control actions vector entries, their range, and dimensions are given in Tables \ref{tab:table1} and \ref{tab:table2}, respectively.
\begin{table}[h]
    \caption{Vehicle State Description}
    \label{tab:table1}
    \centering
    \begin{tabular}{l|c|c|c}
        \textbf{State signal} & \textbf{Symbol} & \textbf{Units} & \textbf{Domain}\\ [0.5ex]
        \hline
        \rule{0pt}{2.5ex}Longitudinal acceleration & $a_x$ & $m\cdot s^{-2}$ & $(-\infty,+\infty)$\\
        \rule{0pt}{2.5ex}Lateral acceleration & $a_y$ & $m\cdot s^{-2}$ & $(-\infty,+\infty)$\\
        \rule{0pt}{2.5ex}Yaw rate & $\dot{\theta}$ & $deg\cdot s^{-1}$ & $(-\infty,+\infty)$\\
        \rule{0pt}{2.5ex}Longitudinal velocity & $v_x$ & $km\cdot h^{-1}$ & $[0, +\infty)$
    \end{tabular}
\end{table}
\begin{table}[h]
    \caption{Control Actions Description}
    \label{tab:table2}
    \centering
    \begin{tabular}{l|c|c|c}
        \textbf{Control} & \textbf{Symbol} & \textbf{Units}  & \textbf{Domain}\\ [0.5ex]
        \hline
        \rule{0pt}{2.5ex}Throttle percentage & $u_t$ & - & $[0,100]$\\
        \rule{0pt}{2.5ex}Brake percentage & $u_b$ & - & $[0,100]$\\
        \rule{0pt}{2.5ex}Steering angle & $u_s$ & $deg$ & $(-180,180)$\\
        \rule{0pt}{2.5ex}Gear & $u_g$ & - & $\{1,2,3,4,5,6\}$
    \end{tabular}
\end{table}


\section{Data Analysis and Preprocessing}
\label{sec:data}
We generate the training data using Assetto Corsa Competizione simulator \cite{AssettoCorsaCompetizione}, paired with MoTeC i2 Pro \cite{MoTeC} to extract channels from the virtual electronic control unit (ECU). This setup is chosen because the data obtained from this simulator are comparable to real-world acquisitions. The software accurately replicates vehicle characteristics, closely aligning with their real-world counterparts. The virtual road surfaces are modelled using Light Detection And Ranging (LiDAR) technology, capturing the details of real track roads, and providing enhanced realism in wheel-ground interaction. In addition, Assetto Corsa Competizione software has become a reference in the literature for obtaining realistic telemetry data \cite{ACC1,ACC2}, also used by professional drivers.

The acquired signals include vehicle states (i.e., accelerometer and speedometer measurements as those indicated in Table \ref{tab:table1}) and driver actions (i.e., throttle and brake percentages, steering wheel angle, and engaged gear as those reported in Table \ref{tab:table2}).

\begin{figure*}[h]
    \centering
    \includegraphics[width=0.9\linewidth]{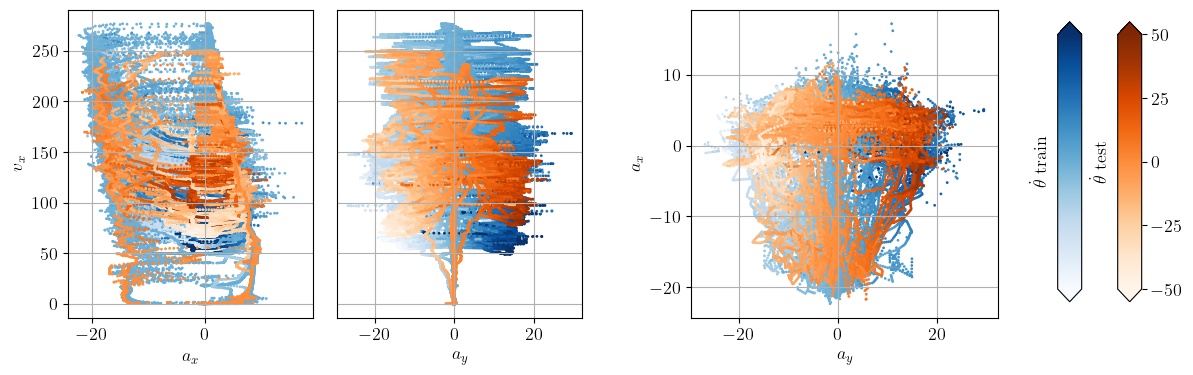}
    \caption{Training and test data for the vehicle state. Orange scale color represent training data, and blue scale color represent test data.}
    \label{fig:dataset}
\end{figure*}

The training set is carefully designed to comprehensively sample a wide range of the vehicle's state space, as shown in Figure \ref{fig:dataset} in blue color. It includes five hot laps—three on the Monza circuit and two on the Valencia circuit—and maneuvers that deviate from the racing line. To better capture the system dynamics under investigation, the recorded laps feature a variety of scenarios, including constant-gear driving, sinusoidal steering inputs at different frequencies, accelerations, and braking at varying gears, intensities, and frequencies. Additionally, the driver autonomously determines the frequencies of the sinusoidal steering inputs and the intensity and timing of acceleration and braking maneuvers, thereby simulating non-ideal scenarios. Relying solely on hot laps where the driver consistently follows the "ideal racing line" would result in a less informative dataset, potentially limiting the model's ability to learn the vehicle's dynamic responses.

The test dataset is constructed on the Misano circuit, shown in Figure 
 \ref{fig:misano} in Appendix \ref{app:tele-relerro}. Similarly to the train data set, it includes a warm-up lap, one hot lap and a lap featuring sinusoidal steering inputs and braking and acceleration at varying intensities. This approach ensures the models are evaluated on a circuit not present in the training set and maneuvers deviating from typical racing scenarios. The test dataset is visualized in Figure \ref{fig:dataset} in orange dots and in Figure \ref{fig:telemetry} in telemetry format in Appendix \ref{app:tele-relerro}.

All acquisitions start and end with the vehicle at a standstill to avoid disrupting the dynamics resulting from the merging process. This approach ensures the proper concatenation of acquisitions without introducing nonphysical ``defects" into the datasets.

The dataset is collected at a sampling frequency of $100~Hz$, under no-wind, sunny weather conditions, with a dry track, uniform tire conditions, constant vehicle load, and fixed setup. For subsequent processing, the data are filtered at different frequencies, yielding four dataset versions: $45\,Hz$, $25\,Hz$, $5\,Hz$, and $0.5\,Hz$, to analyze the effect of noise on model performance in training.
An eighth-order Butterworth filter \cite{butterworth1930} is used to achieve strong selectivity in removing frequencies.

In a race car operating on a track, noise, and vibrations are an inherent part of the process and arise from various factors. These include the suspension components interacting with the road surface, crossing curbs, and transients with significant load transfers that heavily excite the chassis system. Additionally, vibratory phenomena generated by the tires in high-speed contexts should not be overlooked.

\begin{figure}[h]
    \centering
    \includegraphics[width=0.75\linewidth]{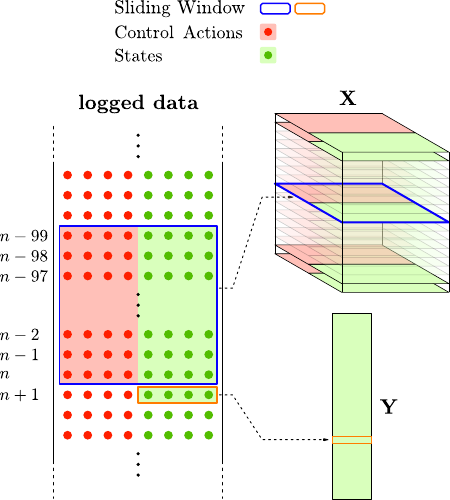}
    \caption{Qualitative representation of the data reshaping process for VeMo architecture. Two sliding windows are used to generate the stack of arrays $\bf X$ and $\bf Y$.}
    \label{fig:data-reshape}
\end{figure}

Before proceeding with the training pipeline, all data are non-dimensionalized based on the maximum values present in the training dataset for the quantities under examination. Table \ref{table:scaling-factors} indicates the reference values used for the non-dimensionalization.
\begin{table}[h]
    \caption{Quantity and Scaling Factors}
    \label{table:scaling-factors}
    \centering
    \begin{tabular}{c|c|c}
        \textbf{Quantity} & \textbf{Scaling Factor} & \textbf{Units} \\[0.5ex]
        \hline
        \rule{0pt}{2.5ex}$u_t$ & 100 & -\\  
        \rule{0pt}{2.5ex}$u_b$ & 100 & -\\  
        \rule{0pt}{2.5ex}$u_s$ & 250 & $deg$\\  
        \rule{0pt}{2.5ex}$u_g$ & 6   & -\\  
        \rule{0pt}{2.5ex}$a_x$ & $2\,g$ & $m\cdot s^{-2}$\\  
        \rule{0pt}{2.5ex}$a_y$ & $2\,g$ & $m\cdot s^{-2}$\\  
        \rule{0pt}{2.5ex}$\dot\theta$ & 60 & $deg\cdot s^{-1}$\\
        \rule{0pt}{2.5ex}$v_x$ & 280 & $km\cdot h^{-1}$\\
    \end{tabular}
\end{table}
The original logged data is organized in a table, with each column representing a signal and each row representing a time step (see logged data in Figure \ref{fig:data-reshape}). We call $T$ the total number of time steps. The logged data is reshaped into two structures compatible with the requirements of the VeMo neural model. Specifically, the input $\mathbf{X}$ is shaped as a three-dimensional array of shape $(T-100,100,8)$, see $\bf X$ in Figure \ref{fig:data-reshape}. The output is shaped as a two-dimensional array of shape $(T-100 \times 4)$, see $\bf Y$ in Figure \ref{fig:data-reshape}. For all analyses, the last second of log data will be used as past states, fixing it at 100 by the sampling rate. The samples used for training and testing are reported in Figures \ref{fig:dataset} for the accelerations, yaw rate, and longitudinal velocity. The data are shown without any filtering. Additional telemetric visualizations for the test data are reported in Appendix \ref{app:tele-relerro} in Figure \ref{fig:telemetry}.


\section{VeMo Model Architecture}
\begin{figure*}[h]
    \centering
    \includegraphics[width=0.8\linewidth, , height=0.3\textheight]{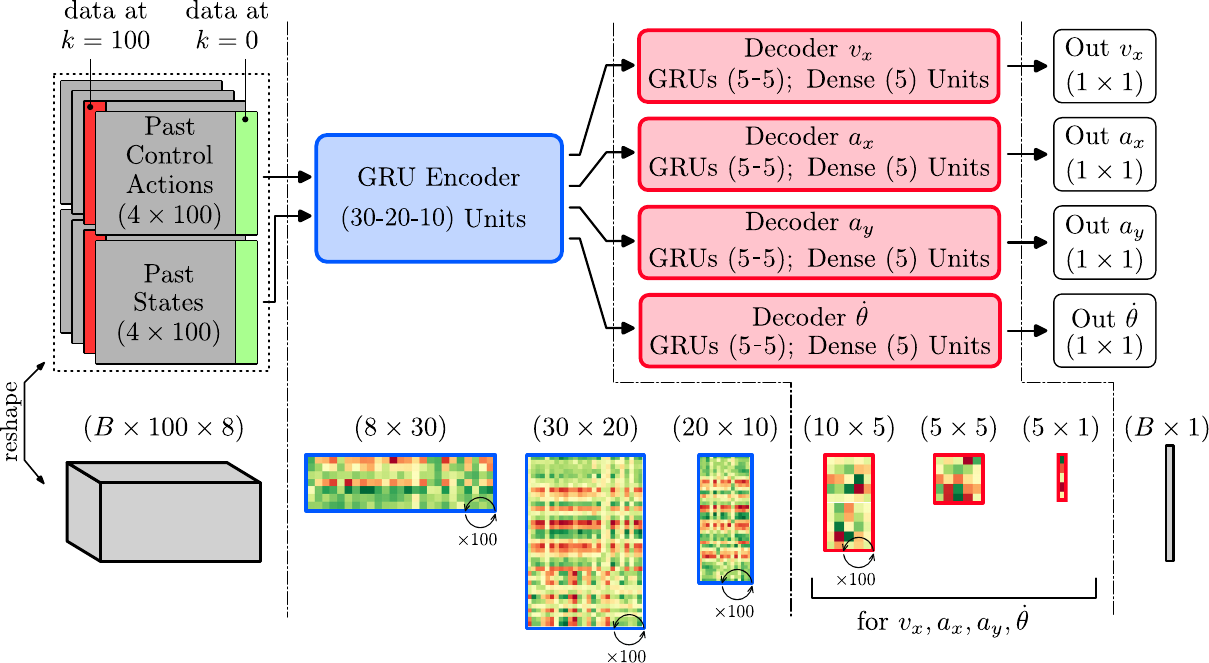}
    \caption{VeMo neural network architecture and a simplified matrix flow, using B as the batch size. The matrices \( W_z \), as described in Appendix \ref{app:gru} are schematically represented.
    The symbol of the curved arrows followed by '$\times100$' indicates that the GRU layer returns the entire sequence of 100 values, while in its absence, it refers only to the last value obtained from the sequential application within the GRU, as shown in Appendix \ref{app:gru}.}
    \label{fig:VeMo}
\end{figure*}

The neural model of VeMo takes the vehicle state vector and its past $k$ states and control actions vector as input. The desired output is the vehicle state vector \( \mathbf{x}_{n+1} \). The model is a neural network based on GRU layers \cite{cho2014rnn}, organized according to an encoder-decoder structure; see Figure \ref{fig:VeMo}.

The encoder section follows a bottleneck approach, using the input to extract a hidden state that captures the coupled dynamics. The decoder section is managed through four main branches, each responsible for decoding the hidden state into their respective target output signals $a_x$, $a_y$, $\dot{\theta}$, and $v_x$.
This approach preserves the dynamic coupling while enabling the branches to specialize in decoding their specific signal. Such approaches are widely known and used, particularly in the field of ML applied to medicine, computer vision, and time series analysis \cite{MultiBranch1,MultiBranch2,MultiBranch3}. Therefore, considering the structure suitable for the task at hand and finding no previous applications of similar architectures in the literature for vehicle dynamics modeling, we proposed it in a simplified form. This choice allows us to limit the model's size, making it applicable in real-world scenarios with low-cost onboard hardware. Since the model is intended to operate alongside control systems, overly large computational models may not be practically feasible for the proposed application. Moreover, since the model is free from physical constraints, it offers greater flexibility in cases where additional input or output channels are required, which can be easily added. Furthermore, the model is not tied to a specific order of differentiation of the state variables. To demonstrate the latter feature, an output is proposed with two accelerations, one angular velocity and one translational.

Attention is also paid to the problem of robustness to noise. To address this aspect, the model is trained using mean absolute error (MAE) loss, and in the encoder section, exponential linear units (ELU) activation functions are employed to take advantage of their properties shown in \cite{elu}. Operating with a structure with separate outputs allows for handling specific losses for each signal, should the need arise, for example, for additional regularization terms.


\section{Experiments \& Results}

\subsection{Experiments}
The proposed model is tested in one-step prediction mode, and four versions with different levels of filtering — $0.5\,Hz$, $5\,Hz$, $25\,Hz$, and $45\,Hz$ — are analyzed. The focus is placed on the noise component because, as is well known, this represents a challenge in supervised machine learning \cite{noise1,noise2,noise3}. For safety-critical applications, such as those proposed for onboard vehicle systems, it is essential to assess how a model, operating with various temporal input sequences, performs under conditions of more or less smoothed data and to evaluate any potential degradation in accuracy.
In this work, to assess the aforementioned performance metric, we train a model for each of the four indicated filters and study its performance against the benchmark filtered at the same cutoff frequency. These models are then provided input data at different filtering levels to evaluate their sensitivity to inputs with higher frequency content than those they have seen during the training phase. In particular, we provide input data containing the following frequencies: $45$, $25$, $15$, $5$, and $1\,Hz$.

\subsection{Evaluation Metrics}
We consider the following metrics to evaluate the model's performance. Where $y_i$ is the ground truth value, $\hat{y}_i$ is the predicted value at the time index $i$, and $N$ is the number of samples.
\begin{itemize}
    \item \textbf{Root Mean Squared Error (RMSE)}
    \[
    RMSE = \sqrt{\frac{1}{N} \sum_{i=1}^{N} (y_i - \hat{y}_i)^2},
    \]
    \item \textbf{Relative Error}
    \[
    \epsilon_{rel} = 100\,\frac{|y_i - \hat{y}_i|}{\underset{i}{\max}(|y_i|)}.\,[\%]
    \] 
    \item \textbf{Maximum Absolute Error}
    \[
    \mathcal{E}_{\text{max}} = \max_{i} \left| y_i - \hat{y}_i \right|.
    \]
\end{itemize}

In Appendix \ref{app:tele-relerro}, Figure \ref{fig:relative_error_time_series}  displays the time evolution of the signal, mean, and median values for the absolute value of $\epsilon_{rel}$.

\subsection{Case with $45\,Hz$ filtering in training}
This section presents the results obtained using a 45 Hz filter. This value is chosen as it is significantly higher than the frequencies typically of interest in vehicle dynamics, which are generally identified in the $0.25\div20\,Hz$ range and referred to as the ``Ride" band \cite{ride1, ride2}.

\begin{figure*}[h]
    \centering
    \includegraphics[width=0.9\textwidth]{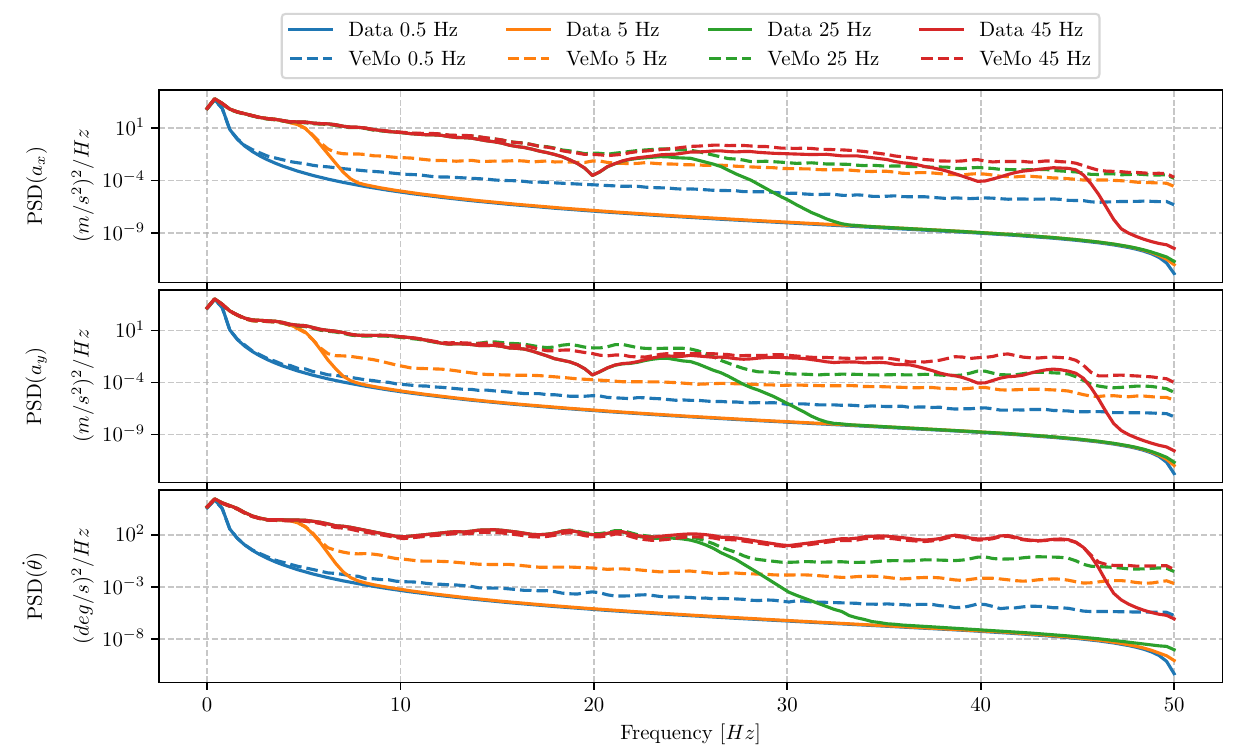}
    \caption{Comparison of the power spectral density of each trained model and the reference data for the frequencies under examination.}
    \label{fig:psd}
\end{figure*}

In Figure \ref{fig:psd}, we plot the power spectral density (PSD) of each trained model and the reference data for the frequencies under examination. For signals where frequency analysis is relevant, we observe that the model accurately captures the frequencies within the range of interest despite the noise in the response. However, the model fails to capture specific spectral features at $20$ and $40\,Hz$ frequencies. Nonetheless, the frequency analysis results are considered satisfactory, as they effectively cover the portion of the spectrum of ``Ride".

\begin{figure*}[h]
    \centering
    \includegraphics[width=0.95\textwidth, height=0.2\textheight]{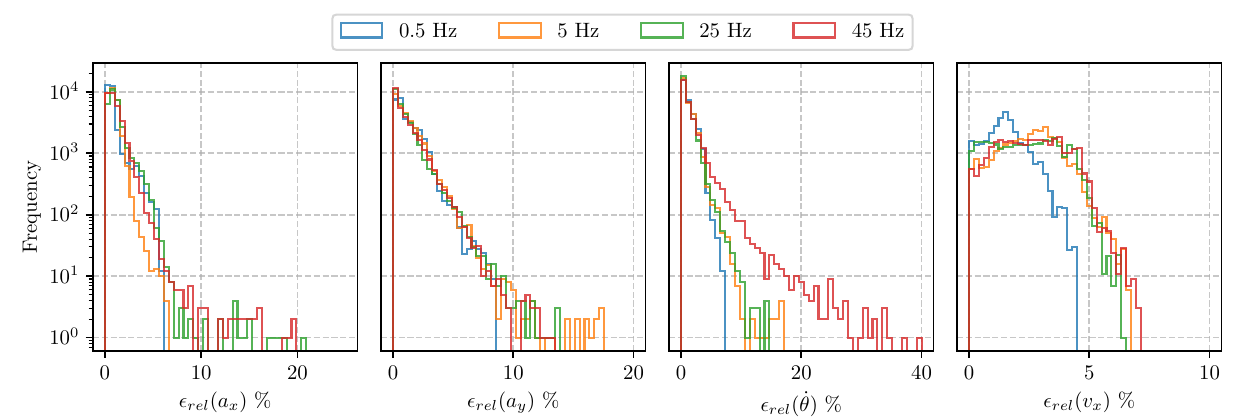}
    \caption{Histogram of the relative error for the three outputs, using VeMo trained with data filtered at $0.5\,Hz$, $5\,Hz$, $25\,Hz$ and $45\,Hz$.}
    \label{fig:hist_aggregated}
\end{figure*}
The analysis of the error histograms indicates that, despite the heavy-tailed distribution, the model consistently predicts the signals, achieving the values in Table \ref{tab:errors45}.
\begin{table}[]
    \caption{Error Metrics for Predicted Signals at $45\,Hz$}
    \label{tab:errors45}
    \centering
    \begin{tabular}{c|c|c|c}
        \textbf{Signal} & \textbf{RMSE} &  \textbf{Mean $\epsilon_{\text{rel}}$ [\%]}  & \textbf{$\mathcal{E}_{\text{max}}$} \\[0.5ex]
        \hline
        \rule{0pt}{2.5ex}        $a_x$          & 0.302  & 1.030 &  4.278 \\
        \rule{0pt}{2.5ex}        $a_y$          & 0.373  & 1.086 &  3.093 \\
        \rule{0pt}{2.5ex}        $\dot{\theta}$ & 1.514  & 1.582 & 22.314 \\
        \rule{0pt}{2.5ex}        $v_x$          & 7.267  & 2.607 &  17.42 \\
    \end{tabular}
\end{table}
As shown in Figure \ref{fig:hist_aggregated} and Table \ref{tab:errors45}, although the mean errors are very low, high values of absolute maximum error emerge. These represent outlier cases due to noise and do not, therefore, pose a problem in prediction.

\subsection{Case with $25\,Hz$ filtering in training}
In this section, we approach the frequencies of the ``Ride" band, significantly narrowing the spectrum of possible frequencies. From this filtering onward, the results exhibit a meaningful spectral content for vehicle dynamics.

In this case, at $25\,Hz$, the obtained spectra exhibit good overlap, as shown in Figure \ref{fig:psd}. As previously observed, despite the noise reduction, the model fails to capture specific features of the signals at $20\,Hz$. Additionally, unlike before, the model's tendency to introduce high-frequency noise into the predicted signal can be observed. Nevertheless, the segment of interest has been correctly modeled.
\begin{table}[h]
    \caption{Error Metrics for Signal Predictions at $25\,Hz$}
    \label{tab:errors25}
    \centering
    \begin{tabular}{c|c|c|c}
        \textbf{Signal} & \textbf{RMSE} &  \textbf{Mean $\epsilon_{\text{rel}}$ [\%]}  & \textbf{$\mathcal{E}_{\text{max}}$} \\[0.5ex]
        \hline
        \rule{0pt}{2.5ex}$a_x$          & 0.331        & 1.168        &  4.494 \\ 
        \rule{0pt}{2.5ex}$a_y$          & 0.361        & 1.055        &  3.185 \\ 
        \rule{0pt}{2.5ex}$\dot{\theta}$ & 0.874        & 1.040        &  8.091 \\ 
        \rule{0pt}{2.5ex}$v_x$          & 6.768        & 2.341        &  15.838 \\ 
    \end{tabular}
\end{table} \noindent
At this cutoff frequency, it can be observed, in Figure \ref{fig:hist_aggregated} and Table \ref{tab:errors25}, that the mean errors are quite consistent with the values obtained for the case at $45\,Hz$. Additionally, the errors due to peaks remain, negatively impacting the maximum absolute error observed.
Despite the high $\mathcal{E}_{\text{max}}$, the RMSE and mean relative error remain low across all signals, implying that the model performs well in the majority of cases. This behavior suggests the peaks are anomalies rather than representative of general behavior.

The results obtained by providing the model trained at $25\,Hz$ with input signals at $45\,Hz$ are presented in Table \ref{tab:errors25wnoise} to demonstrate how the model reacts to an input with more noise than encountered during training.
\begin{table}[h]
    \caption{Error Metrics for Inputs at $45\,Hz$ using $25\,Hz$ model}
    \label{tab:errors25wnoise}
    \centering
    \begin{tabular}{c|c|c|c}
        \textbf{Signal} & \textbf{RMSE} &  \textbf{Mean $\epsilon_{\text{rel}}$ [\%]}  & \textbf{$\mathcal{E}_{\text{max}}$} \\[0.5ex]
        \hline
        \rule{0pt}{2.5ex}$a_x$          & 0.346       & 1.201        &  4.375 \\ 
        \rule{0pt}{2.5ex}$a_y$          & 0.412        & 1.164        &  4.957 \\ 
        \rule{0pt}{2.5ex}$\dot{\theta}$ & 1.823        & 1.954        &  28.736 \\ 
        \rule{0pt}{2.5ex}$v_x$          & 6.633        & 2.276        &  15.523 \\ 
    \end{tabular}
\end{table}\noindent
The input of a noisier signal does not show significant variations in the model's performance, indicating that the model obtained is sufficiently robust to noise.

\subsection{Case with $5\,Hz$ filtering in training}
In this case, we position ourselves in the lowest frequency quartile of the band of interest. Here, the signal can be considered smooth for the processes under examination, making it an ideal case.

This case study, Figure \ref{fig:psd}, demonstrates how the model accurately captures the entire spectrum within the $5\,Hz$ range for the three analyzed signals. Here, too, we observe an imperfect attenuation of high frequencies.

For the error histograms in Figure \ref{fig:hist_aggregated}, we observe that although the heavy-tail shape remains, the outliers are significantly reduced. Moreover, the metrics in Table \ref{tab:errors5} indicate an improvement in performance in this case, highlighting the impact of noise in the context of machine learning and the necessity of exploring it for applications of this nature. It is also noted that the maximum absolute error is remarkably lower than in the previous two cases. At these frequencies, the model can accurately capture all the significant peaks and is no longer affected by the negative impact of high-frequency crests.
\begin{table}[H]
    \caption{Error Metrics for Predicted Signals at $5\,Hz$}
    \label{tab:errors5}
    \centering
    \begin{tabular}{c|c|c|c}
        \textbf{Signal} & \textbf{RMSE} &  \textbf{Mean $\epsilon_{\text{rel}}$ [\%]}  & \textbf{$\mathcal{E}_{\text{max}}$} \\[0.5ex]
        \hline
        \rule{0pt}{2.5ex}$a_x$          & 0.209  & 0.848 & 1.305 \\ 
        \rule{0pt}{2.5ex}$a_y$          & 0.388  & 1.213 & 3.893 \\ 
        \rule{0pt}{2.5ex}$\dot{\theta}$ & 0.899  & 1.084 & 9.425 \\ 
        \rule{0pt}{2.5ex}$v_x$          & 7.022  & 2.561 & 16.368 \\ 
    \end{tabular}
\end{table}\noindent
For this case, we also show in Table \ref{tab:errors5wnoise} how the model trained for the frequency of the scenario under examination behaves when inputs with richer frequency content are provided. Specifically, we report the results for cases where inputs at $45$, $25$, and $15\,Hz$ are fed into the model trained at a frequency of $5\,Hz$.
\begin{table}[h]
    \caption{Error Metrics for Inputs at Different Frequencies using training data filtered at $5\,Hz$ model}
    \label{tab:errors5wnoise}
    \centering
    \begin{tabular}{c|c|c|c|c}
        \textbf{Signal} & \textbf{Metric} & \textbf{45 Hz} & \textbf{25 Hz} &\textbf{15 Hz} \\ [0.5ex]
        \hline \rule{0pt}{2.5ex}
        $a_x$ & RMSE & 0.562  & 0.558  & 0.547 \\ \rule{0pt}{2.5ex}
        $a_y$ & RMSE & 0.526  & 0.518  & 0.497 \\ \rule{0pt}{2.5ex}
        $\dot{\theta}$ & RMSE & 1.617  & 1.584  & 1.434\\
        \rule{0pt}{2.5ex}
        $v_x$ & RMSE & 6.802  & 6.849  & 6.905 \\ [0.5ex]
        \hline \rule{0pt}{2.5ex}
        $a_x$ & mean $\epsilon_{\text{rel}}$ & 1.647 & 1.637 & 1.614 \\ \rule{0pt}{2.5ex}
        $a_y$ & mean $\epsilon_{\text{rel}}$ & 1.506 & 1.489 & 1.436 \\ \rule{0pt}{2.5ex}
        $\dot{\theta}$ & mean $\epsilon_{\text{rel}}$ & 1.820 & 1.785 & 1.632 \\ 
        \rule{0pt}{2.5ex}
        $v_x$ & mean $\epsilon_{\text{rel}}$ & 2.464  & 2.489  & 2.514 \\ [0.5ex]
        \hline \rule{0pt}{2.5ex}
        $a_x$ & $\mathcal{E}_{\text{max}}$  & 10.968  & 10.733  & 10.485  \\ \rule{0pt}{2.5ex}
        $a_y$ & $\mathcal{E}_{\text{max}}$  & 5.391  & 5.257  & 5.185  \\ \rule{0pt}{2.5ex}
        $\dot{\theta}$ & $\mathcal{E}_{\text{max}}$ & 17.022  & 15.346 & 14.183 \\
        \rule{0pt}{2.5ex}
        $v_x$ & $\mathcal{E}_{\text{max}}$ & 34.470  & 34.028  & 31.841 \\
    \end{tabular}
\end{table}\noindent
The results in \ref{tab:errors5wnoise} show that the model at $5\,Hz$ is generally robust to noise, with only occasional spikes appearing in the predicted signals, without altering the physical consistency of the other values in the signals, thus explaining the very high maximum error values obtained.
Only a counterintuitive trend is observed for the longitudinal velocity value, which shows a slight deterioration as it approaches the cutoff frequency experienced during training.

\subsection{Case with $0.5\,Hz$ filtering in training}
We effectively remove much of the signal of interest for this final evaluation at $0.5\,Hz$. Therefore, it serves as a valuable case for completing the sensitivity analysis. Still, it represents a signal of limited relevance for the dynamics in question, constituting a case of significantly simplified dynamics.
\begin{table}[h]
    \caption{Error Metrics for Predicted Signals at $0.5\,Hz$}
    \label{tab: errors05}
    \centering
    \begin{tabular}{c|c|c|c}
        \textbf{Signal} & \textbf{RMSE} &  \textbf{Mean $\epsilon_{\text{rel}}$ [\%]}  & \textbf{$\mathcal{E}_{\text{max}}$} \\[0.5ex]
        \hline
        \rule{0pt}{2.5ex}$a_x$          & 0.239 & 0.841 & 1.117 \\ 
        \rule{0pt}{2.5ex}$a_y$          & 0.351 & 1.228 & 1.759 \\ 
        \rule{0pt}{2.5ex}$\dot{\theta}$ & 0.714 & 1.130 & 3.043 \\
        \rule{0pt}{2.5ex}$v_x$          & 4.169 & 1.494 & 11.052 \\
    \end{tabular}
\end{table}
\noindent
From the values shown in Figure \ref{fig:hist_aggregated} and Table \ref{tab: errors05}, we observe how, in this last case, the metrics are slightly further improved, especially in terms of maximum absolute error, which in this condition is free from any unwanted spikes.

\begin{figure}
    \centering
    \includegraphics[height=0.22\textheight]{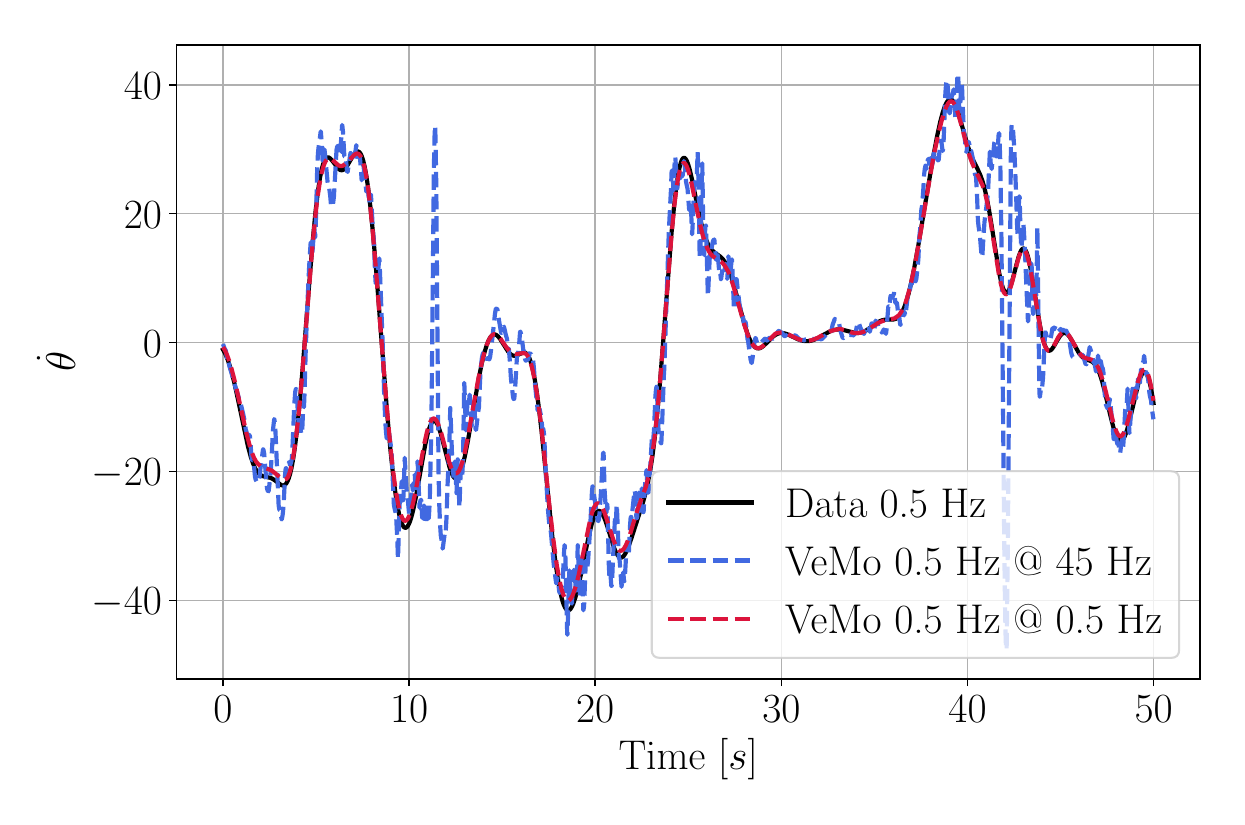}
    \caption{Comparison of the target \(\dot{\theta}\) at $0.5\,Hz$ and the output of the VeMo model trained using a cutoff frequency of $0.5\,Hz$ and fed respectively with a signal filtered at $45\,Hz$ in blue and filtered at $0.5\,Hz$ in red.}
    \label{fig:noise on yr}
\end{figure}

As for all previous cases, the results obtained by providing inputs filtered with higher cutoff frequencies, specifically $45$, $25$, $15$, $5$, and $1\,Hz$, are reported below in Table \ref{tab:errors05wnoise}.
\begin{table}[]
    \caption{Error Metrics for Inputs at Different Frequencies using training data filtered at $0.5\,Hz$}
    \label{tab:errors05wnoise}
    \centering
    \begin{tabular}{c|c|c|c|c|c|c}
        \textbf{Signal} & \textbf{Metric}    & \textbf{45 Hz}        & \textbf{25 Hz}        & \textbf{15 Hz}        & \textbf{5 Hz}         & \textbf{1 Hz}       \\ [0.5ex]
        \hline
        \rule{0pt}{2.5ex}$a_x$  & RMSE          & 1.898 & 1.898  & 1.897 & 1.862  & 1.309  \\
        \rule{0pt}{2.5ex}$a_y$  & RMSE          & 1.877  & 1.877  & 1.878 & 1.867  & 1.359  \\
        \rule{0pt}{2.5ex}$\dot{\theta}$ & RMSE  & 4.858 & 4.858  & 4.858 & 4.818  & 3.166  \\
        \rule{0pt}{2.5ex}$v_x$  & RMSE          & 3.867 & 3.860  & 3.863 & 3.786  & 2.977  \\
        \hline
        \rule{0pt}{2.5ex}$a_x$ & mean $\epsilon_{\text{rel}}$ & 5.989 & 5.989 & 5.988 & 5.895 & 4.538 \\
        \rule{0pt}{2.5ex}$a_y$ & mean $\epsilon_{\text{rel}}$ & 5.549 & 5.549 & 5.548 & 5.502 & 4.505 \\
        \rule{0pt}{2.5ex}$\dot{\theta}$ & mean $\epsilon_{\text{rel}}$ & 6.318 & 6.318 & 6.317 & 6.242 & 4.505 \\
        \rule{0pt}{2.5ex}$v_x$ & mean $\epsilon_{\text{rel}}$ & 1.046 & 1.045 & 1.046 & 1.045 & 0.908 \\
        \hline
        \rule{0pt}{2.5ex}$a_x$ & $\mathcal{E}_{\text{max}}$  & 12.380  & 12.478  & 12.465  & 11.047  & 5.818  \\
        \rule{0pt}{2.5ex}$a_y$ & $\mathcal{E}_{\text{max}}$  & 19.612 & 19.582 & 19.544 & 19.401 & 9.624  \\
        \rule{0pt}{2.5ex}$\dot{\theta}$ & $\mathcal{E}_{\text{max}}$  & 55.42 & 55.380 & 55.386 & 55.144 & 23.932 \\
        \rule{0pt}{2.5ex}$v_x$ & $\mathcal{E}_{\text{max}}$  & 24.944  & 24.936  & 24.932  & 24.490  & 18.034  \\
    \end{tabular}
\end{table}\noindent
Despite the significant noise introduced into the model, the results undergo limited degradation, returning acceptable average metrics. The maximum errors are instead caused by the same phenomenon observed in the previous case, where sharp peaks appear in the predicted signal, as shown in Figure \ref{fig:noise on yr} on \(\dot{\theta}\) signal. Furthermore, in this case, there is no longer a worsening effect on the longitudinal velocity metrics as the filter used during training is approached.

\subsection{Results} 
In this section, we assess the results from the experiments detailed earlier. The findings indicate that the $5\,Hz$ model achieves very low error rates and demonstrates physical consistency. Additionally, our analysis of the power spectral density (PSD) of the predicted signals shows a favorable match with the reference data. This implies that the model effectively captures essential characteristics in the frequency domain, maintaining accuracy across various frequency components. Moreover, the models display notable resilience to noise levels exceeding those encountered during training. Specifically, in the worst-case scenario where a $45\,Hz$ signal is processed by a model trained at $0.5\,Hz$, the maximum degradation observed in the yaw rate, $\dot{\theta}$, reaches $4.144$ $deg/s$ as computed by the RMSE metric.

Among the different state components, the longitudinal speed shows the most significant error, with a tendency to increase as speed rises, likely due to the test scenarios' high variability and demanding nature of the circuit. The previously considered test at Misano involves strong speed gradients, making them particularly challenging for any model.  

To further assess how velocity error behaves in a low longitudinal gradients and high-speed condition, a case typical of oval circuit racing, we design a specific evaluation scenario. Furthermore, this context is a testing ground for autonomous driving in racing, from several years, through the Indy Autonomous Challenge \cite{iac}. Therefore, it is considered of interest to the community. 

\begin{figure}
    \centering
    \includegraphics[height=0.25\textheight]{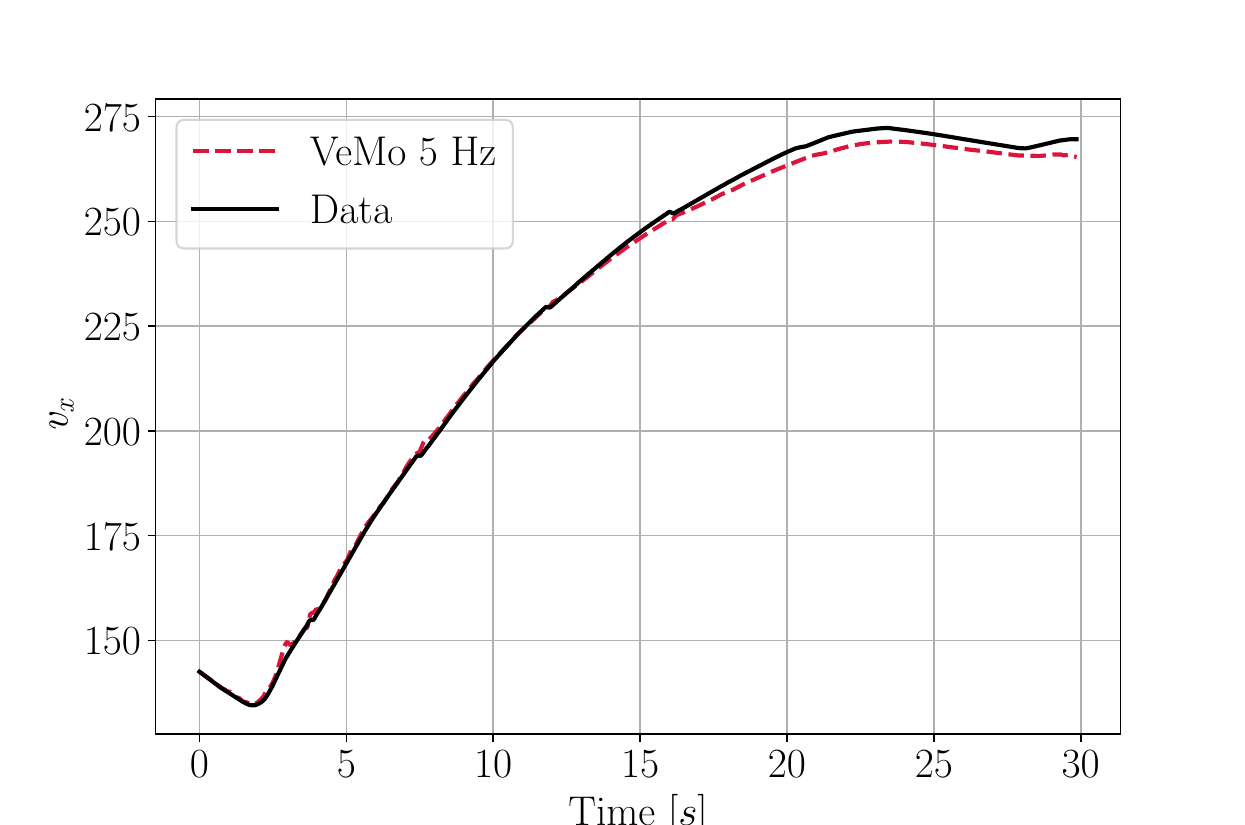}
    \caption{VeMo comparison on $v_x$ stabilized maneuver. 
    }
    \label{fig:vx_stabi}
\end{figure}

We construct a benchmark that reflects the maneuvers achievable in an oval circuit, reproducing a driving sequence characterized by an exit from a turn with high lateral acceleration, followed by a phase of reaching and stabilizing speed around $270$ $km/h$, as shown in Figure \ref{fig:vx_stabi}. We remove the initial offset to examine how the error in the velocity component signal $v_x$ evolves with speed. This choice yields zero error at the first dataset point and allows us to evaluate only the error trend.
In this setting, for the $v_x$ state component, we report an RMSE of $1.943$ $km/h$, a mean $\epsilon_{\text{rel}}$ of $0.575\%$, and a maximum error $\mathcal{E}_{\text{max}}$ of $4.240$ $km/h$ due only to deviation effects. These results highlight how previously reported errors were linked to a significantly more challenging scenario, reinforcing the suitability of VeMo for engineering applications.

\section{Conclusions}
In this work, we addressed the problem of modeling vehicle dynamics in contexts where little to no information is available about the vehicle system under examination. These scenarios are often associated with autonomous or assisted driving sectors, as third parties typically develop these applications on an existing vehicle.

We propose VeMo, a neural model that utilizes an encoder-decoder architecture featuring a shared encoder to create dynamic coupling, along with a dedicated decoder for each output. This model takes into account longitudinal and lateral accelerations, yaw rate, and longitudinal velocity as past states, along with the control actions taken by the driver, to predict the next state of the vehicle. Four different cases are studied to evaluate the effect of noise on the model, each with a different filtering level, specifically at $45$, $25$, $5$, and $0.5\,Hz$.

The results demonstrate that the VeMo model can effectively represent signals, even when the training dataset includes noisy data. This indicates that the models maintain their robustness against noise, showing only a slight decrease in performance despite being exposed to signals with higher frequency content than those encountered during training.

VeMo shows satisfactory performance at $5\,Hz$, achieving mean relative errors of $0.848\%$, $1.213\%$, $1.084\%$, and $2.561\%$ for longitudinal acceleration, lateral acceleration, yaw rate, and longitudinal velocity signals, respectively. Notably, these results were accomplished without imposing any physical constraints.

These results show that fully data-driven approaches can work well without simplified physical constraints. By augmenting the state vector and control inputs with onboard measurements, the model can be adapted to other vehicle platforms, such as aerial or tracked systems.

We are optimistic about the results obtained from this initial version of the neural network architecture. We believe that future developments can enhance the overall performance of the system, providing a tool with application in areas such as Advanced Driver Assistance Systems, Automated Vehicles and Vehicle Control.

\section*{Credit authorship contribution statement}

Girolamo Oddo: conceptualization (equal); investigation (equal); methodology (equal); software (lead); validation (lead); visualization (supporting); writing - original draft (equal); writing - review and editing (equal).\\

Roberto Nuca: conceptualization (equal); investigation (equal); methodology (equal); software (supporting), visualization (lead); writing - original draft (equal); writing - review and editing (equal); supervision (equal).\\

Matteo Parsani: conceptualization (equal); methodology (supporting); funding acquisition (lead); resources (lead); writing - review and editing (equal); supervision (equal); project administration (lead).

\section*{Author declarations}
The authors have no conflicts of interest to disclose.

\section*{Acknowledgments}
This work was funded by King Abdullah University of Science \& Technology (grant numbers: BAS/1/1663-01-01 and REP/1/6262-01-01). Roberto Nuca is a member of GNCS/INdAM.
We thank Mr. Giuseppe Bisogno for supporting the data generation activities using Assetto Corsa Competizione.


\appendix
\subsection{GRU Ovierview}
\label{app:gru}
The complete structure of the gated recurrent unit (GRU) layer is described below. Initially, for $t=0$, the output vector is $h_0=0$.
\begin{align*}
    z_t       & = \sigma(W_z x_t + U_z h_{t-1} + b_z) \\
    r_t       & = \sigma(W_r x_t + U_r h_{t-1} + b_r) \\
    \hat{h}_t & = \phi(W_h x_t + U_h (r_t \odot h_{t-1}) + b_h) \\
    h_t       & = (1 - z_t) \odot h_{t-1} + z_t \odot \hat{h}_t
\end{align*}

The symbol $\odot$ denotes the Hadamard product (element-wise product between vectors), $d$ denotes the number of input features and $f$ the number of output features. The Table \ref{tab:gru_params} summarizes the variables involved in the GRU architecture.

\begin{table}[h!]
    \centering
    \caption{Gate Recurrent Unit Variables}
    \label{tab:gru_params}
    \begin{tabular}{c|l|l}
        \textbf{Vector} & \textbf{Space} & \textbf{Description} \\[0.5ex]
        \hline 
        \rule{0pt}{2.5ex}
        $x_t$       & $\mathbb{R}^d$        & Input \\ \rule{0pt}{2.5ex}
        $h_t$       & $\mathbb{R}^f$        & Output \\
        \rule{0pt}{2.5ex}
        $\hat{h}_t$ & $\mathbb{R}^f$        & Candidate activation \\ \rule{0pt}{2.5ex}
        $z_t$       & $(0,1)^f$             & Update gate \\ \rule{0pt}{2.5ex}
        $r_t$       & $(0,1)^f$             & Reset gate \\ \rule{0pt}{2.5ex}
        $W$         & $\mathbb{R}^{f \times d}$ & Learned parameter \\ \rule{0pt}{2.5ex}
        $U$         & $\mathbb{R}^{f \times f}$ & Learned parameter \\ \rule{0pt}{2.5ex}
        $b$         & $\mathbb{R}^f$        & Learned parameter \\
    \end{tabular}
\end{table}
\noindent
The activation functions are defined as
\begin{align*}
    \sigma(x) &= \frac{1}{1 + e^{-x}} &&\text{logistic function,} \\
    \phi(x)   &= \tanh(x)             &&\text{hyperbolic tangent.}
\end{align*}

This is the architecture proposed in the original implementation \cite{cho2014rnn}. The VeMo model proposed in this work adopts $\sigma(x)= \text{ELU}(x)$ (exponential linear unit) for the encoder. This choice is justified by its better robustness to noise as shown in \cite{elu}. The $\text{ELU}$ function is defined as follows:
\[
\text{ELU}(x) =
\begin{cases}
x & \text{if } x > 0, \\
\alpha\,(e^x - 1) & \text{if } x \leq 0.
\end{cases}
\]
As shown in the equations above, the GRU model involves several matrices and vectors learned during the training process. A brief overview is reported below for a better understanding:
\begin{itemize}
    \item \( W_z \), $W_r$, $W_h$: weight matrices transforming the input vector \( x_t \) into the hidden state space. Each of these matrices has dimensions \( f \times d \), where \( f \) is the number of output features (hidden state size) and \( d \) is the number of input features.

    \item \( U_z \), \( U_r \), \( U_h \): weight matrices transforming the previous hidden state \( h_{t-1} \) into the hidden state space. Each of these matrices has dimensions \( f \times f \).

    \item \( b_z \), \( b_r \), \( b_h \): bias vectors added to linear transformations. Each of these vectors has dimensions \( f \). These latter are optional and, specifically, were not used for the VeMo architecture since the required output signals do not need static offsets, as they are zero when the system is at rest.

    \item \( z_t \): update gate vector determining how much of the past information (from \( h_{t-1} \)) to keep and how much of the new information (from \( \hat{h}_t \)) to add.

    \item \( r_t \): reset gate vector determining how much of the past information to forget.

    \item \( \hat{h}_t \): the candidate activation vector is a combination of the current input \( x_t \) and the reset-gated previous hidden state \( h_{t-1} \).

    \item \( h_t \): the new hidden state is a combination of the previous hidden state \( h_{t-1} \) and the candidate activation vector \( \hat{h}_t \), modulated by the update gate \( z_t \).
\end{itemize}

\subsection{Test Data and Relative Error as Time Series}\label{app:tele-relerro}
\begin{figure}
    \centering
    \includegraphics[height=0.15\textheight]{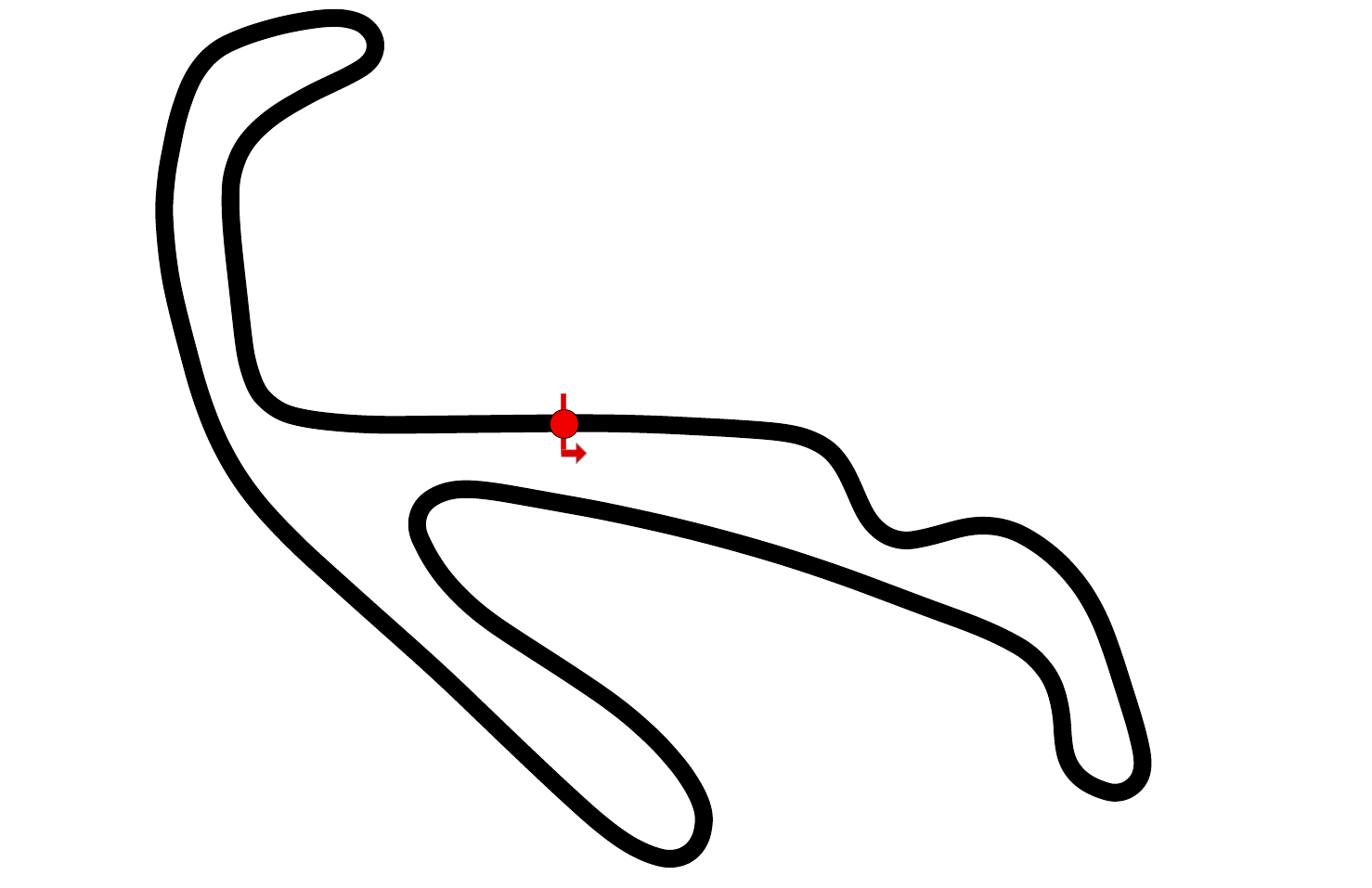}
    \caption{Test Circuit: Misano World Circuit ``Marco Simoncelli".}
    \label{fig:misano}
\end{figure}

\begin{figure*}[h]
    \centering
    \includegraphics[angle=0,width=0.9\linewidth]{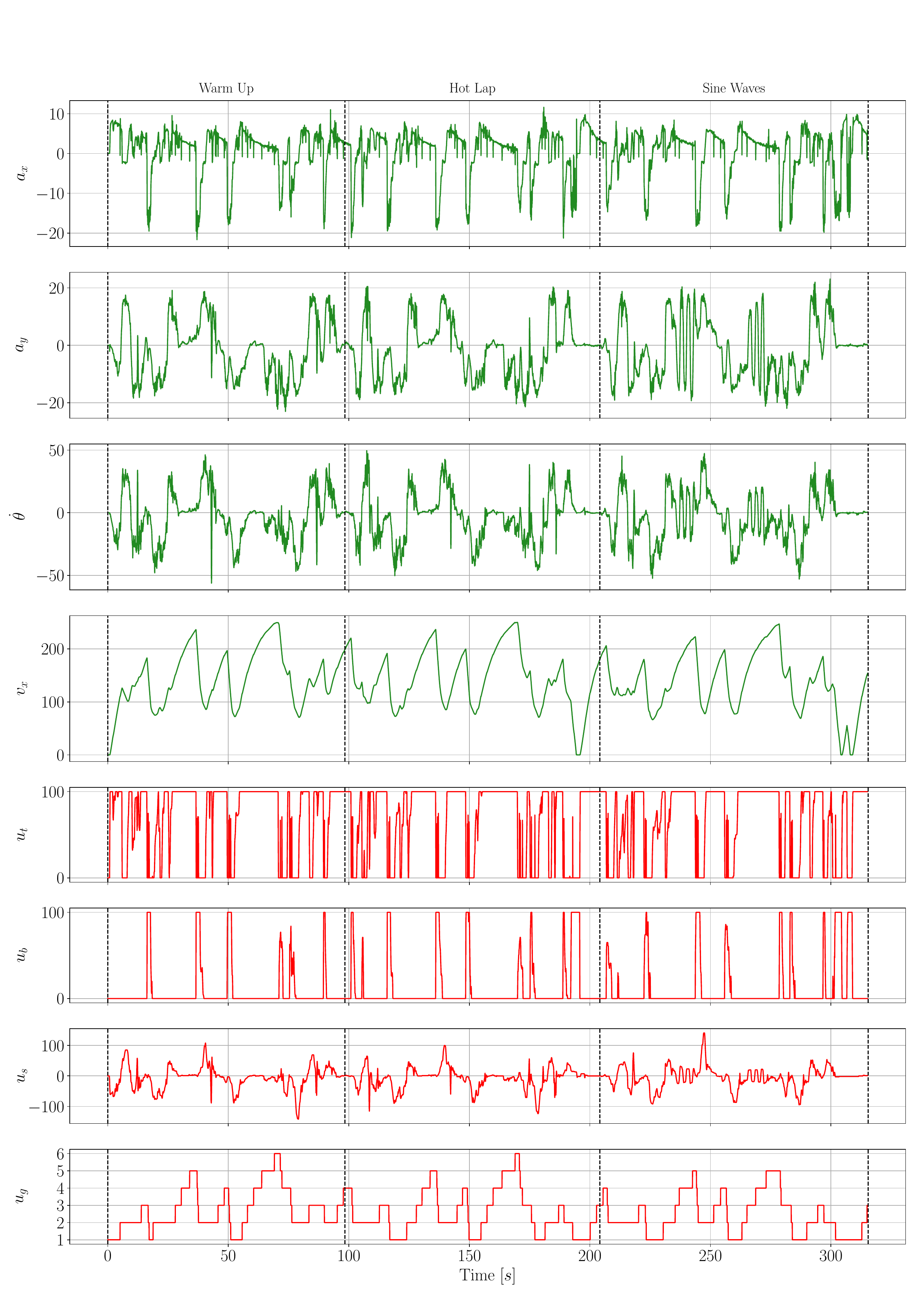}
    \caption{Time series of the states and action controls measured by the telemetry.}
    \label{fig:telemetry}
\end{figure*}

\begin{figure*}[h]
    \centering
    \includegraphics[angle=-90,width=0.9\linewidth]{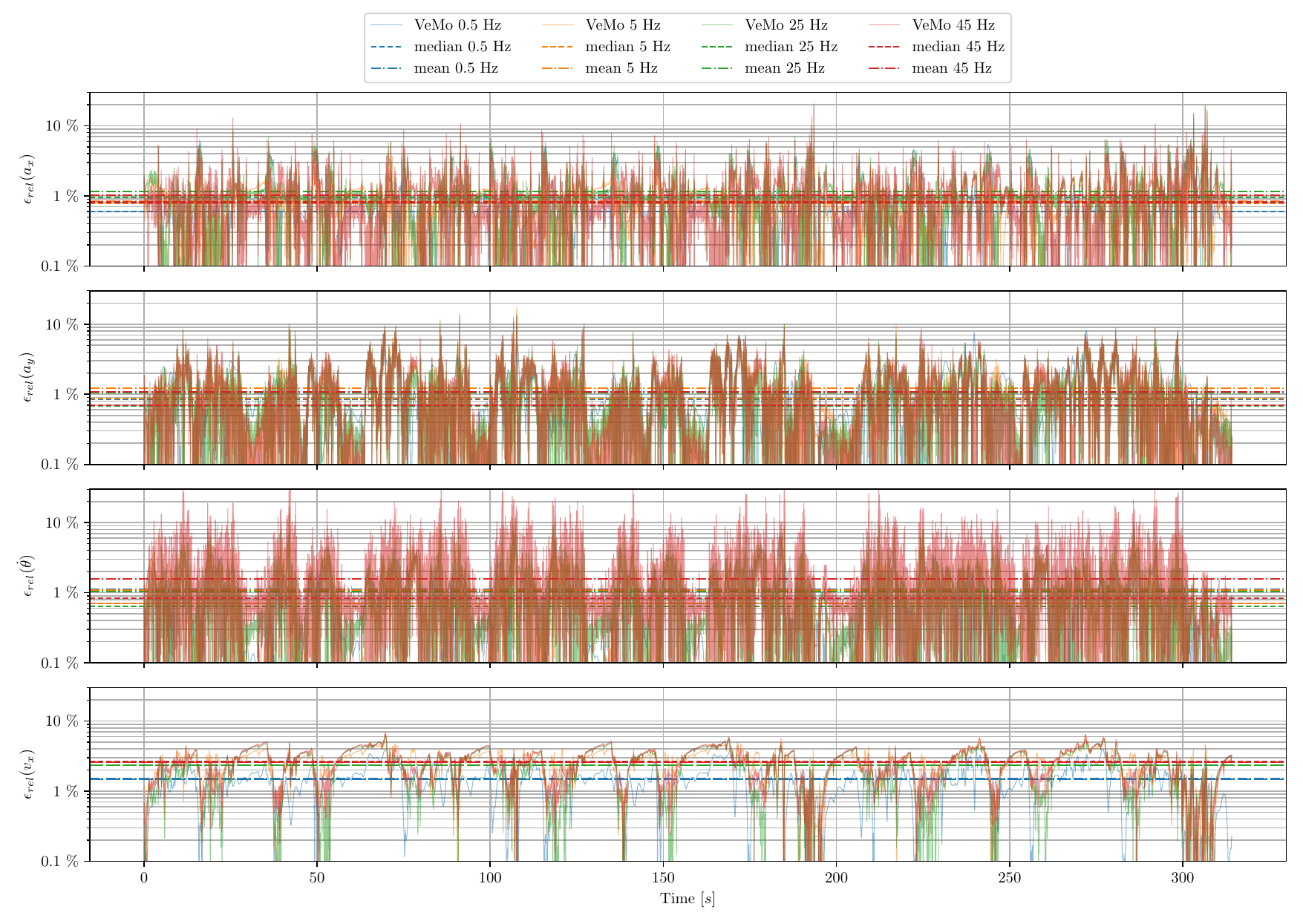}
    \caption{Temporal signal of the relative error for the three outputs using VeMo trained with data filtered at $0.5~Hz$, $5~Hz$, $25~Hz$, and $45~Hz$. Horizontal lines indicate the mean and median of the displayed signal. Different colors indicate different filtering frequencies.}
    \label{fig:relative_error_time_series}
\end{figure*}

The Misano circuit and data used for testing the VeMo model are presented as unfiltered time series in Figures \ref{fig:misano}, \ref{fig:telemetry}. These time series, accompanied by the time history of the relative error shown in Figure \ref{fig:relative_error_time_series}, provide the full picture with data of the performance assessment of the VeMo model. Specifically, in Figure \ref{fig:relative_error_time_series}, we report the history of the relative error for the vehicle state vector components, i.e., the longitudinal acceleration, $a_x$, the lateral acceleration, $a_y$, the yaw rate $\dot\theta$ and longitudinal velocity $v_x$. Four families of lines are shown for the four frequencies analyzed in this work.

In addition, the visual view of the data in Figure \ref{fig:relative_error_time_series} allows a better understanding of how control actions influence the vehicle states.



\end{document}